\def\year{2020}\relax
\documentclass[letterpaper]{article} 
\usepackage{aaai20}  
\usepackage{times}  
\usepackage{helvet} 
\usepackage{courier}  
\usepackage[hyphens]{url}  
\usepackage{graphicx} 
\usepackage{graphicx}  
\usepackage{amsmath}
\usepackage{amsfonts}
\usepackage{booktabs}

\title{Learning Light Field Angular Super-Resolution via a Geometry-Aware Network}
\author{Jing Jin,\textsuperscript{\rm 1}\protect\thanks{This work was supported in part by the Hong Kong RGC Early Career Scheme under Grant 9048123 (CityU 21211518), and in part by the Huawei Innovative Research Program under Grant 9231332. Corresponding author: Junhui Hou (jh.hou@cityu.edu.hk)}
Junhui Hou,\textsuperscript{\rm 1}
Hui Yuan,\textsuperscript{\rm 2}
Sam Kwong\textsuperscript{\rm 1}\\
\textsuperscript{\rm 1}City University of Hong Kong,
\textsuperscript{\rm 2}Shandong University
}

\begin{document}
\maketitle

\begin{abstract}
The acquisition of light field images with high angular resolution is costly. Although many methods have been proposed to improve the angular resolution of a sparsely-sampled light field, they always focus on the light field with a small baseline, which is captured by a consumer light field camera.  By making full use of the intrinsic \textit{geometry} information of light fields, in this paper we propose an end-to-end learning-based approach aiming at angularly super-resolving a sparsely-sampled light field with a large baseline. Our model consists of two learnable modules and a physically-based module. Specifically, it includes a depth estimation module for explicitly modeling the scene geometry, a physically-based warping for novel views synthesis, and a light field blending module specifically designed for light field reconstruction. Moreover, we introduce a novel loss function to promote the preservation of the light field parallax structure. Experimental results over various light field datasets including large baseline light field images demonstrate the significant superiority of our method when compared with state-of-the-art ones, i.e.,  our method improves the PSNR of the second best method up to 2 dB in average, while saves the execution time 48$\times$. In addition, our method preserves the light field parallax structure better.
\end{abstract}

\section{Introduction} 
\noindent
Light field images provide rich information of 3D scenes by recording not only the intensity but also the direction of light rays. Conventional light field acquisition methods include camera array \cite{lf2005array} and computer-controlled gantry \cite{lfgantry}, which sample the light field at different viewpoints through single or multiple exposures. Due to the increase of hardware complexity, it is very costly to obtain high angular resolution using these systems. Recently, commercial light field cameras \cite{lytro,raytrix} attract a lot of attention because of their portability. However, the limitation of sensor resolution leads to an inevitable trade-off between the spatial and angular resolution of the captured light field images.

        \begin{figure}[!t]
        \centering
        \includegraphics[width=0.8\columnwidth]{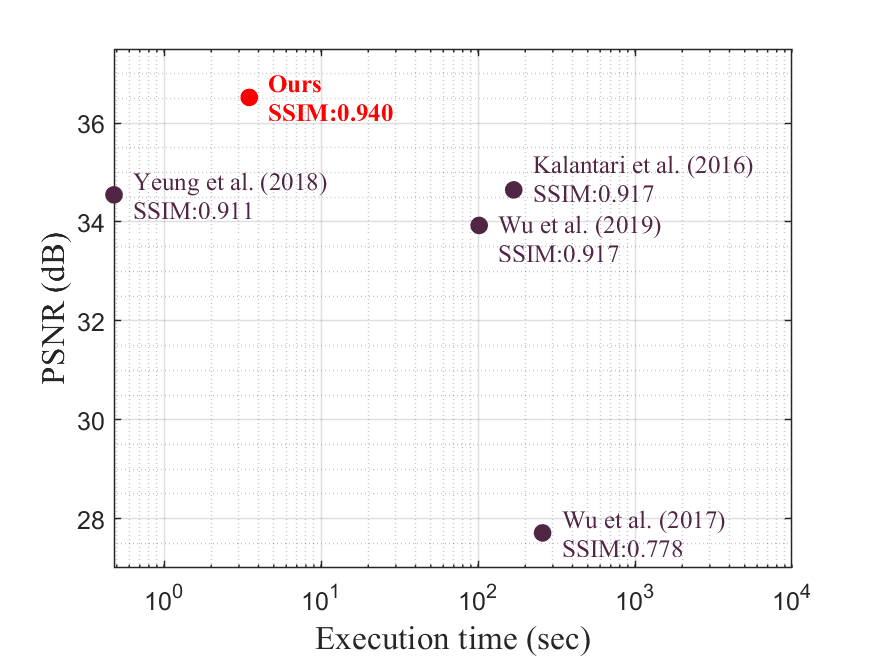}
        \caption{Comparisons of the execution time (in second) and reconstruction quality (PSNR/SSIM) of different methods. Here, a sparse light field containing $2\times2$ views of spatial resolution $512\times512$ is super-resolved to a high angular resolution light field containing $7\times7$ views. The PSNR/SSIM value refers to the average over 48 light fields with a disparity range of $[-4,4]$. Our method produces the highest reconstruction quality while takes less time than all other methods except one.
        }
        \label{fig_timevspsnr}
        \end{figure}

        \begin{figure*}[!t]
        \centering
        \includegraphics[width=0.8\textwidth]{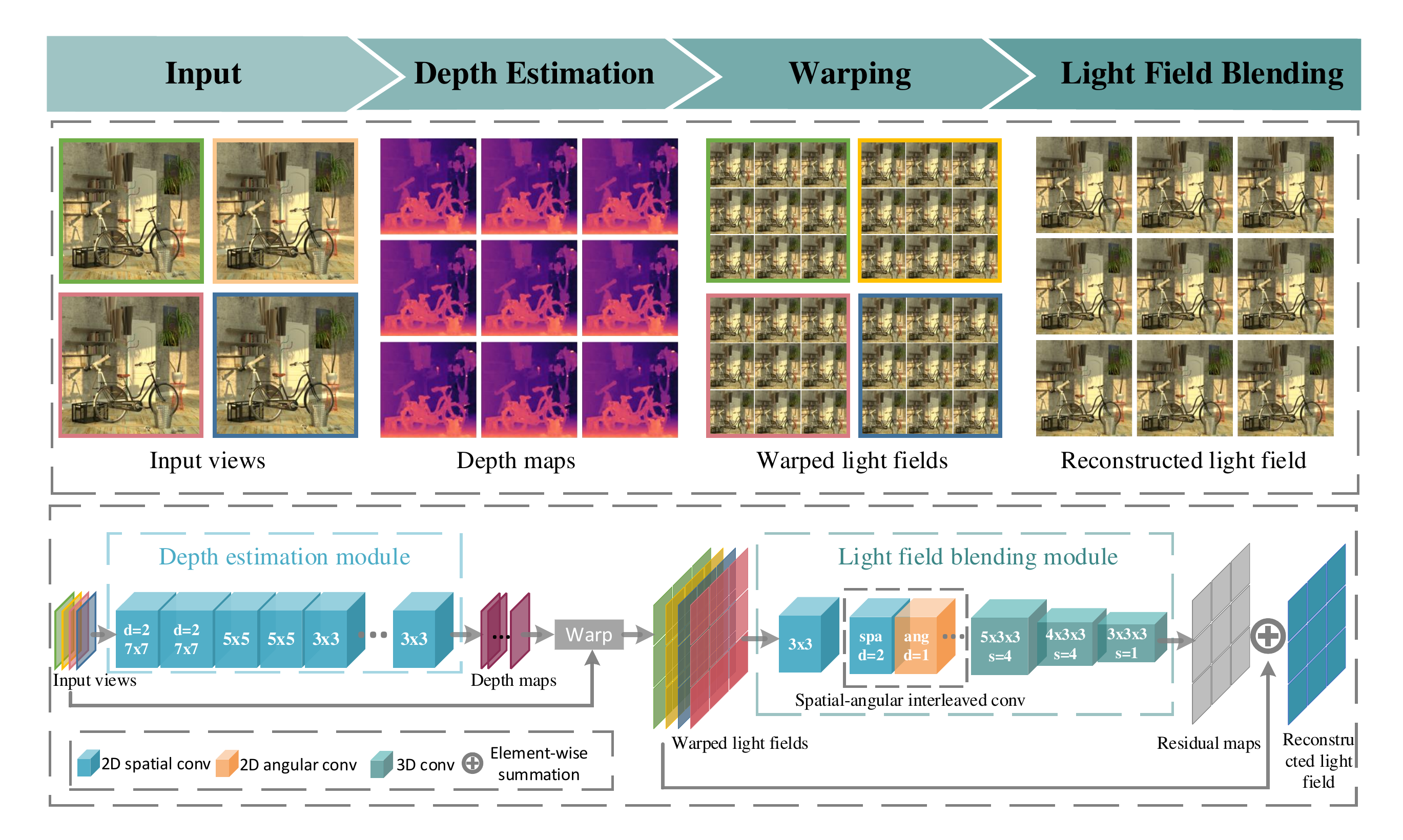}
        \caption{The flowchart of the proposed method for light field angular super-resolution. The reconstruction of a $3\times3$ light field from a $2\times2$ sparse one is depicted for demonstration. An output with higher angular resolution can be easily realised using a similar architecture. Three modules are involved in our method. The depth estimation module predicts a depth map for each view of the high angular resolution light field, the warping module initially generates the novel views by physically warping the input views based on the estimated depth maps, and the light field blending module explores the spatial-angular relations among the warped light fields (i.e., the light field parallax geometry) to reconstruct a high angular resolution light field.
        }
        \label{fig_workflow}
        \end{figure*}

To mitigate this problem, many studies have been devoted to the light field angular super-resolution. Particularly, inspired by the great success of convolutional neural networks (CNNs) \cite{krizhevsky2012classification,sisr2014srcnn}, many learning-based methods have been proposed to enable light field angular super-resolution from an extremely small set of views \cite{lfrec2016kalantari,lfrec2018Yeung,lfrec2018pseudo,lfrec2017wu:blur,lfrec2019wu:shear}. They always focus on the reconstruction of light fields captured by commercial light field cameras, where the baseline of the input views is very small. These methods can be roughly classified into two categories: non-depth based and depth based. When the baseline of the input views increases, methods without modeling the scene depth \cite{lfrec2018Yeung,lfrec2018pseudo,lfrec2017wu:blur} always produce obvious artifacts in the synthesized novel views. Although utilizing depth information makes it easier to handle inputs with large disparities, existing depth-based methods cannot achieve acceptable performance for large-baseline sampling yet, because they either neglect the angular relations between the reconstructed views \cite{lfrec2016kalantari} or underuse the spatial information of the input views \cite{lfrec2019wu:shear}. 

In view of these issues, in this paper we focus on the angular super-resolution of light field images with a large baseline, and propose an end-to-end trainable method, by making full use of the intrinsic geometry information of light fields. As illustrated in Fig. \ref{fig_workflow}, our method consists of three modules. Specifically,  we first estimate a 4D depth map for the high angular resolution light field, which provides a depth for each light ray in the 4D light field.
Compared with the direct prediction of the intensity of each light ray, the estimation of depth maps could be much more accurate. The resulting 4D depth is then utilized to synthesize all novel views by backward warping. For the blending module which attempts to fuse the warped images, different from existing methods which perform the fusion of the warped images for each view independently via multiple 2D convolutional layers \cite{lfrec2016kalantari}, we adopt a light field blending instead. That is, the blending considers not only the complementarity between images warped from different input views, but also the angular correlations between warped images at different novel views, as shown in Fig. \ref{fig_analysis_blend}. Furthermore, to improve the ability of preserving the light field parallax structure, we introduce a novel loss based on the gradient of epipolar-plane images (EPIs). This loss can be potentially used in other light field related tasks.

We demonstrate the advantage of our method on the angular super-resolution of a $2\times 2$ light field  to  a $7\times 7$ one by using various datasets containing light fields with relatively large disparities. That is, as shown in Fig. \ref{fig_timevspsnr}, our method is able to reconstruct a high angular resolution light field with higher quality both qualitatively and quantitatively. Moreover, our method is very efficient compared with the state-of-the-art ones.

\section{Related Work}
The problem of light field angular super-resolution has been studied for decades. Existing methods can be roughly classified into two categories: non-depth based methods and depth-based methods.

For non-depth based methods, various priors for light field images were used to solve the inverse problem of super-resolution, such as a mixture of Gaussians, sparisty and low-rank \cite{lfrec2008bayesian,lfrec2014sparsity,lfrec2016tensor,lfrec2012gmm,lfrec2018shearlet}. These methods always require large number of input views. Based on compressive sensing principles, light field images with a large amount of data can be recovered from fewer acquisitions \cite{lfrec2012compressive,lfrec2013compressive,lfrec2017compressiveDeep}. However, the input views need to be sampled in specific patterns, which increases the difficulty of acquisition.

Recently, some methods using CNNs have been proposed. Yoon et al. \shortcite{lfrec2015yoon} proposed an end-to-end network to first improve the spatial resolution of each view individually, then generate novel views one by one based on neighboring input views.
The performance of this method is very limited, as the relations between input views are not explored. More recently, different methods have been proposed to explore the regular structure of the light field.
Wu et al. \shortcite{lfrec2017wu:blur} applied CNNs to reconstruct 2D EPIs. Similarly, Wang et al. \shortcite{lfrec2018pseudo} proposed to process 3D volumes of the stacked EPIs. These methods are not able to fully exploit the 4D information of the light field yet. Yeung et al. \shortcite{lfrec2018Yeung} proposed to process the 4D data using pseudo 4D filters, i.e. spatial-angular separable filters, which produces good results on real-world images captured by light field cameras.

Depth-based methods for light field angular super-resolution
typically first estimate depth map at the novel view or the input view, and then use it to synthesize the novel view by backward- or forward-warping \cite{lfdepth2014wanner,lfdepth2015jeon}. The performance of these methods is heavily relied on the accuracy of estimated depth maps. Recently, this pipeline was modeled using CNNs \cite{lfrec2016kalantari}, which consists of depth and color estimation components. This method struggles against inputs with large baseline as the depth estimation component fails to capture the long-distance correspondences. Moreover, this method independently synthesizes novel views while neglects their inter-view correlations. Wu et al. \shortcite{lfrec2019wu:shear} also proposed a learning-based method leveraging the depth information. They computed the depth value based on the structure of sheared EPIs, and upsampled the EPIs for light field angular super-resolution. As an EPI is a 2D slice of a 4D light field, the EPI-based method cannot utilize the information of spatial context, making it difficult to handle complicated scenes.
Layered representations are also modeled using CNNs for novel view synthesis \cite{viewsyn2018mpi,viewsyn2019localfusion}, which is able to generate novel views at different positions using single representation.

\section{The Proposed Method}
Let $L(\mathbf{x},\mathbf{u})$ denote a 4D high angular resolution light field, where $\mathbf{x}=(x,y)$ is the spatial coordinate and $\mathbf{u}=(u,v)$ is the angular coordinate, and $L(\mathbf{x},\mathbf{u'})$ be a small set of views belonging to $L$, where $\mathbf{u'}$ is the angular position sampled at the $(u,v)$ grid. Our objective is to super-resolve $L(\mathbf{x},\mathbf{u'})$ in the angular domain to construct a high angular resolution light field denoted as $L(\mathbf{x},\mathbf{u'})$, which is as close as to $L(\mathbf{x},\mathbf{u})$. This problem can be formulated as:
        \begin{equation}
          \widehat{L}(\mathbf{x},\mathbf{u}) =  f(L(\mathbf{x},\mathbf{u'})),
        \end{equation}
where $f$ is the function representing the angular super-resolution process to be learned.

To reconstruct a high angular resolution light field from sparse views, the intensities of unsampled light rays are required to be predicted.
A naive method is straightforwardly applying deep CNNs to regress the values. It relies on the powerful representation ability of deep CNNs to learn the light field image statistics from a large variety of data.
However, when the baseline of the input views increases, the ghosting and blurry effects are severe because local convolutions always have trouble in modeling long-distance relations. 

\textit{Remark}: One unique characteristic of the light field is the intrinsic geometry information, i.e., the geometry relation among the involved views (or the light field parallax structure), and likewise the geometry of captured scenes/objects. It is expected that the performance of angular super-resolution will be enhanced by fully exploring such valuable geometry information.  

To this end, our proposed method consists of three modules, i.e., depth estimation $f_d$, warping $f_w$ and light field blending $f_b$.
Specifically, we first estimate a depth map for each view in the light field, which indicates the correlations between the known light rays to unknown ones. Based on the 4D ray depth, novel views can be initially generated by warping the input views, giving a set of warped light fields. The warped images inevitably have distortions due to depth estimation errors,  non-Lambertian regions, and occlusions.
Different from commonly used blending method, which combines the images warped from different views to individually produce a novel view, we propose a light field blending strategy, which explores the angular relations among the warped light fields to  preserve the geometry structure of the reconstructed light field.

\textbf{Depth estimation.}
In this module, a 4D ray depth, denoted as $D(\mathbf{x},\mathbf{u})$, is estimated from the input views:
        \begin{equation}
          D(\mathbf{x},\mathbf{u}) =  f_d(L(\mathbf{x},\mathbf{u'})).
        \end{equation}
The estimation of the 4D depth map from sparse views is based on the regular structure of a light field, called light field parallax structure, which can be formulated as:
        \begin{equation}
          L(\mathbf{x},\mathbf{u}) =  L(\mathbf{x}+d\Delta\mathbf{x},\mathbf{u}+\Delta\mathbf{u}),
        \end{equation}
where $d$ is the depth for point $L(\mathbf{x},\mathbf{u})$. Based on this property, we believe a sequential of convolutional layers is able to learn the 4D depth map. Note that we do not use the ground-truth depth maps as supervision, and  the depth estimation is completely induced by the following warping module.

The architecture of the depth estimation module is depicted in Fig. \ref{fig_workflow}. The network consists of nine layers of convolution, each followed by a ReLU activation layer except the last one. To find the correspondences between input views with large disparities for depth estimation, the network is required to have sufficient receptive field.
Therefore, We use $7\times7$ kernels with a dilation rate of 2 for the first two layers. Then the kernel size is decreased to $5\times5$ and $3\times3$ in the rest layers. Such a setting provides a receptive field of 43, which is sufficient for input views with a  disparity range of $[-21.5,21.5]$.

\textbf{Warping.}
Based on the estimated depth maps, novel views can be synthesized by warping the input views. The warping can be formulated as:
        \begin{equation}
        \begin{aligned}
            W(\mathbf{x},\mathbf{u},\mathbf{u'}) =& f_w(L(\mathbf{x},\mathbf{u'}),D(\mathbf{x},\mathbf{u}))\\
            =& L(\mathbf{x}+D(\mathbf{x},\mathbf{u})(\mathbf{u}-\mathbf{u'}),\mathbf{u'}),    
        \end{aligned}
        \end{equation}
where $W(\mathbf{x},\mathbf{u},\mathbf{u'}) $ denotes a novel view at angular position $\mathbf{u}$ produced by warping an input view at $\mathbf{u'}$. 

The reconstruction errors of the warped light fields are minimized to provide proper instruction for the depth estimation network as the ground truth depth maps are not available in practice.
Moreover, the smoothness of each depth map is encouraged by penalizing the spatial gradient.
Finally, the training loss for depth estimation module is formulated as:
        \begin{equation}
           \ell_d = \sum_{\mathbf{x},\mathbf{u}} \left( \sum_{\mathbf{u'}} \left| L(\mathbf{x},\mathbf{u}) - W(\mathbf{x},\mathbf{u},\mathbf{u'})\right| +\nabla_{\mathbf{x}}D(\mathbf{x},\mathbf{u})  \right).
        \end{equation}

        \begin{figure}[!t]
        \centering
        \includegraphics[width=0.8\columnwidth]{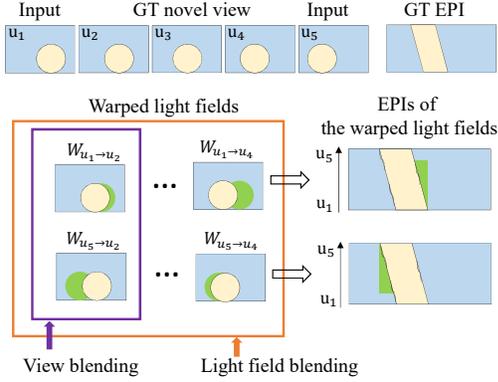}
        \caption{Illustration of two different blending strategies: view blending vs. light field blending. In this example, $\mathbf{u}_1, \cdots, \mathbf{u}_5$ denote 5 views of the ground truth light field. Suppose we need to reconstruct views $\mathbf{u}_2,\cdots,\mathbf{u}_4$ from input views $\mathbf{u}_1$ and $\mathbf{u}_5$ by blending the warped images. $W_{\mathbf{u}_i\rightarrow \mathbf{u}_j}$ stands for the resulting image by warping $\mathbf{u}_i$ to the location of $\mathbf{u}_j$. Green regions are the occluded regions and always have errors after warping. The distortion of the edges in the EPIs of the warped light fields is caused by inaccurate depth estimation. To synthesize $\mathbf{u}_k$ ($k=2, 3, 4$), \textit{view blending} only blends $W_{\mathbf{u}_1\rightarrow \mathbf{u}_k}$ and $W_{\mathbf{u}_5\rightarrow \mathbf{u}_k}$.
        In contrast, our proposed \textit{light field blending} employs all 6 warped images from the 2 input views to the other 3 views, which is able to take advantage of the spatial-angular relations between \textit{warped light fields} to recover the geometry structure in the EPIs.  
        }
        \label{fig_analysis_blend}
        \end{figure}
        
\textbf{Light field blending.}
The light fields initially warped from input views inevitably contain distortions due to two reasons. First, the depth estimation module is not able to predict the ray depth accurately, especially on challenging areas such as textureless regions and repeat patterns. This problem is difficult to solve especially without the ground truth depth maps as supervision. Second, even with ground truth depth maps, the warping operator will introduce errors in occluded regions as no source pixels can be found in the input views \cite{flow2018occlusion}.
As a results, the linear geometry structures in EPIs of the warped light fields could be distorted, and errors could appear in the occluded regions, as shown in Fig. \ref{fig_analysis_blend}.

Existing methods produce the final reconstruction by blending the images warped from different input views using sequential 2D spatial convolutional layers \cite{lfrec2016kalantari}. In this paper, we call this strategy as \textit{view blending}. View blending is not suitable for light field reconstruction here, as the linear  geometry structures of the EPIs are not taken into consideration. To this end, we propose a novel blending strategy, called \textit{light field blending}. The core idea is exploring the angular relations between warped views to recover the geometry structure of the EPIs. Fig. \ref{fig_analysis_blend} is a toy example to show the difference between view blending and light field blending. We use 3D light field for simplification, which can be easily extended to 4D light field.

The light field blending is implemented using a deep CNN. The network architecture is depicted in Fig. \ref{fig_workflow}. Suppose the size of the high angular resolution light field and the input sparse light field are $(H\times W\times M\times N)$ and $(H\times W\times M'\times N')$, respectively, then the warped light fields $W$ has a size of $(H\times W \times MN\times M'N' )$. We first extract $64$ feature maps from the $M'N'$ warped images for each novel view individually. Next, to explore the relations between views of a light field, we adapt the interleaved spatial-angular convolutions \cite{sepfilter2017video,lfsr2019sas,lfrec2018Yeung}.
That is, sequential 2D convolutional layers are alternatively applied on the spatial and angular dimension, which enables to fully explore the directional relations between spatial patches while needs fewer computational resources compared with 4D convolutions. To increase the receptive field in spatial dimension, dilation is used in the spatial convolution. Following this spatial-angular feature extraction, three layers of 3D strided convolution are used to reconstruct the residual map. Finally, a light field image is reconstructed as:
        \begin{equation}
        \begin{aligned}
            \widehat{L}(\mathbf{x},\mathbf{u}) = W(\mathbf{x},\mathbf{u},\mathbf{u'_1})+f_b(W(\mathbf{x},\mathbf{u},\mathbf{u'})),
        \end{aligned}
        \end{equation}
where $W(\mathbf{x},\mathbf{u},\mathbf{u'_1})$ is the light field warped from the first one of the input views.

The light field blending network is supervised by minimizing the reconstruction error of the predicted light field $\widehat{L}$:
        \begin{equation}
        \ell_b =  \sum_{\mathbf{x},\mathbf{u}}  \left| L(\mathbf{x},\mathbf{u}) - \widehat{L}(\mathbf{x},\mathbf{u})  \right|.
        \end{equation}

        \begin{table*}[!t]
        \caption{Quantitative comparisons (PSNR/SSIM) of different methods  over \textit{HCI} dataset.}
        \label{table_hci}
        \centering
        \resizebox{0.9\textwidth}{!}{
        \begin{tabular}{c | c | c c c c | c }
        \toprule[2pt]
        Light field & Disparity range & Wu et al. \shortcite{lfrec2017wu:blur} & Wu et al. \shortcite{lfrec2019wu:shear} & Yeung et al. \shortcite{lfrec2018Yeung} & Kalantari et al. \shortcite{lfrec2016kalantari} &  Ours\\
        \midrule[1pt]
        \textit{bedroom} & $[ -1.7, 2.2 ]$ & 30.06/0.809 & 39.15/0.961 & 38.22/0.957 & 38.77/0.959 & \textbf{41.98}/\textbf{0.975}\\
        \textit{bicycle} & $[ -1.7, 1.7 ]$ & 26.17/0.762 & 30.84/0.924 & 32.92/0.945 & 32.37/0.935 & \textbf{34.03}/\textbf{0.954} \\
        \textit{herbs}   & $[ -3.1, 1.8 ]$ & 26.86/0.694 & 30.80/0.831 & 31.05/0.836 & 31.70/0.847 & \textbf{32.76}/\textbf{0.882}\\
        \textit{dishes}  & $[ -3.1, 3.5 ]$ & 23.46/0.710 & 26.59/0.876 & 27.00/0.863 & 28.56/0.893 & \textbf{29.63}/\textbf{0.938}\\
        \midrule[1pt]
        \multicolumn{2}{c|}{Avg. over 4 light fields}& 26.64/0.744 & 31.84/0.898 & 32.30/0.900 & 32.85/0.909 & \textbf{34.60}/\textbf{0.937}\\
        \bottomrule[2pt]
        \end{tabular}
        }
        \end{table*}

        \begin{table*}[!t]
        \caption{Quantitative comparisons (PSNR/SSIM) of different methods over \textit{HCI old} dataset.}
        \label{table_hci_old}
        \centering
        \resizebox{0.9\textwidth}{!}{
        \begin{tabular}{c | c | c c c c | c }
        \toprule[2pt]
        Light field & Disparity range & Wu et al. \shortcite{lfrec2017wu:blur} & Wu et al. \shortcite{lfrec2019wu:shear} & Yeung et al. \shortcite{lfrec2018Yeung} & Kalantari et al. \shortcite{lfrec2016kalantari} &  Ours\\
        \midrule[1pt]
        \textit{buddha}    & $[-0.85,1.54]$ & 32.86/0.916 & 42.91/0.986 & 44.03/0.988 & 42.47/0.985 & \textbf{45.65}/\textbf{0.991}\\
        \textit{buddha2}   & $[-0.70,1.20]$ & 32.63/0.902 & 38.03/0.966 & 40.61/0.973 & 39.51/0.969 & \textbf{41.48}/\textbf{0.975}\\
        \textit{stillLife} & $[-2.71,2.56]$ & 21.64/0.550 & 24.63/0.792 & 24.14/0.771 & 24.78/0.797 & \textbf{25.67}/\textbf{0.854}\\
        \textit{papillon}  & $[-1.17,0.89]$ & 34.55/0.936 & 41.42/0.981 & 44.73/0.986 & 43.04/0.983 & \textbf{45.51}/\textbf{0.987}\\
        \textit{monasroom} & $[-0.79,0.72]$ & 35.45/0.946 & 41.06/0.983 & 44.92/0.989 & 43.09/0.985 & \textbf{45.88}/\textbf{0.990}\\
        \midrule[1pt]
        \multicolumn{2}{c|}{Avg. over 5 light fields}& 31.43/0.850  & 37.61/0.942 & 39.69/0.941 & 38.58/0.944 & \textbf{40.84}/\textbf{0.960}\\	
        \bottomrule[2pt]
        \end{tabular}
        }
        \end{table*}

        \begin{table*}[!t]
        \caption{Quantitative comparisons (PSNR/SSIM) of different methods  over \textit{Inria DLFD} dataset. 4 light fields were selected to show individual results.}
        \label{table_inria}
        \centering
        \resizebox{0.9\textwidth}{!}{
        \begin{tabular}{c | c | c c c c | c }
        \toprule[2pt]
        Light field & Disparity range & Wu et al. \shortcite{lfrec2017wu:blur} & Wu et al. \shortcite{lfrec2019wu:shear} & Yeung et al. \shortcite{lfrec2018Yeung} & Kalantari et al. \shortcite{lfrec2016kalantari} &  Ours\\
        \midrule[1pt]
        \textit{Black$\&$white}& $[-1.62,0.10]$ & 21.77/0.600 & 33.73/0.969 & 29.31/0.923 & 30.62/0.925 & \textbf{34.69}/\textbf{0.974}\\
        \textit{Furniture}     & $[-2.06,1.92]$ & 28.35/0.852 & 36.62/0.949 & 38.36/0.948 & 36.73/0.935 & \textbf{40.62}/\textbf{0.962}\\
        \textit{Three pillows} & $[-2.33,1.98]$ & 21.15/0.534 & 24.48/0.809 & 24.50/0.755 & 24.64/0.807 & \textbf{25.88}/\textbf{0.917}\\
        \textit{White roses}   & $[-1.52,3.38]$ & 25.25/0.719 & 35.60/0.962 & 36.27/0.960 & 36.28/0.960 & \textbf{40.59}/\textbf{0.981}\\
        \midrule[1pt]
        \multicolumn{2}{c|}{Avg. over 39 light fields}& 25.05/0.740 & 32.35/0.911 & 31.65/0.892 & 32.53/0.899  & \textbf{34.83}/\textbf{0.933}\\	
        \bottomrule[2pt]
        \end{tabular}
        }
        \end{table*}

\textbf{EPI gradient loss.}
To further preserve the valuable light field parallax structure, i.e. promote the geometry consistency between the reconstructed novel views, we propose a novel loss function based on the gradient of EPIs.

An EPI is a 2D slice of the 4D light field, which
can be constructed by fixing one dimension of the spatial and angular domain, respectively. The horizontal and vertical EPI can be represented as $E_{y^\ast,v^\ast}(x,u)=L(x,y^\ast,u,v^\ast)$ and $E_{x^\ast,u^\ast}(y,v)=L(x^\ast,y,u^\ast,v)$, respectively. Due to the regular and symmetric distribution of views in a light field, an EPI is composed of linear geometry structures, and the slope of the lines indicate the depth of corresponding scene points. Therefore, EPIs of the reconstructed light field provide straightforward evaluation for the light field structure.

Our proposed EPI gradient loss is defined as the $\ell_1$ distance between the gradient of EPIs constructed from the predicted and the ground truth light field:
        \begin{equation}
        \begin{aligned}
            \ell_e =  \sum_{y,v} & ( \left|  \nabla_x E_{y,v}(x,u)- \nabla_x\widehat{E}_{y,v}(x,u)  \right|\\
            +& \left|  \nabla_u E_{y,v}(x,u)- \nabla_u\widehat{E}_{y,v}(x,u)  \right|  )   \\
            + \sum_{x,u} & ( \left|  \nabla_y E_{x,u}(y,v)- \nabla_y\widehat{E}_{x,u}(y,v)  \right|\\
            +& \left|  \nabla_v E_{x,u}(y,v)- \nabla_v\widehat{E}_{x,u}(y,v)  \right|  ).
        \end{aligned}
        \end{equation}

\textbf{Training details.}
The final objective of our whole network is: 
        \begin{equation}
            min \quad \ell_d +  \ell_b + \lambda\ell_e,
        \end{equation}
where $\lambda$ is the weighting for the EPI gradient loss. Our model is trained to predict a light field with $7\times7$ views from four corner views. The dataset used for training consists of 20 scenes from \textit{HCI} dataset \cite{lfdataset2016hci}. All images have the spatial resolution of $512\times512$, and the disparity range of $[-4,4]$.

During training, each image was randomly and spatially cropped into $96\times96$ patches. To keep the spatial resolution unchanged, padding of zeros was used for all convolutional layers. The model was implemented with PyTorch. We used Adam optimizer \cite{kingma2014adam} with $\beta_1=0.9$ and $\beta_2=0.999$. The learning rate was set to $1e^{-4}$ initially and decreased by a factor of $0.5$ every $5e^3$ epochs. The codes are available at \url{https://github.com/jingjin25/LFASR-geometry}.

        \begin{figure*}[!t]
        \centering
        \includegraphics[width=0.88\textwidth]{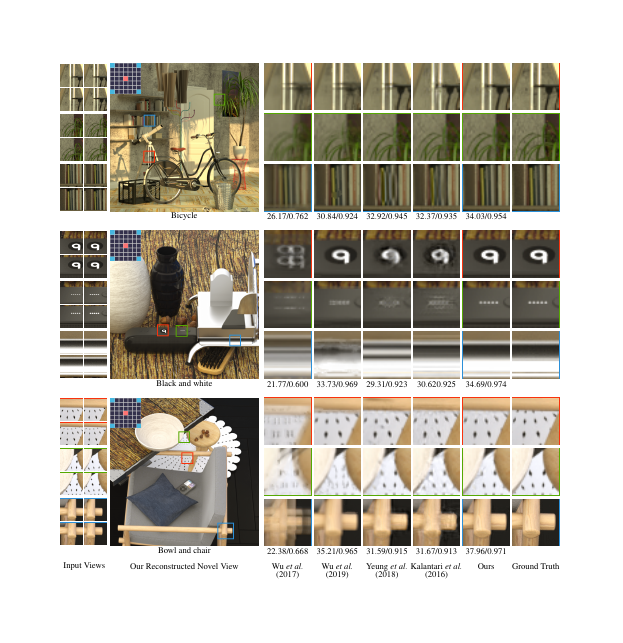}
        \caption{Visual comparisons of different methods on the reconstructed center view from four corner input views.}
        \label{fig_visual_patch}
        \end{figure*}

\section{Experimental Results}
We compared with four state-of-the-art learning-based methods primarily developed for light field angular super-resolution, including Wu et al. \shortcite{lfrec2017wu:blur}, Wu et al. \shortcite{lfrec2019wu:shear}, Yeung et al. \shortcite{lfrec2018Yeung} and Kalantari et al. \shortcite{lfrec2016kalantari}. All these models except Wu et al. \shortcite{lfrec2017wu:blur} with training codes available were re-trained using the same dataset and the suggested training configurations by the authors for fair comparisons.

To evaluate the performance of different methods on inputs with large baselines, 3 datasets containing totally 48 light fields with a disparity range of $[-4,4]$ were used, namely, \textit{HCI} \cite{lfdataset2016hci}, \textit{HCI old} \cite{lfdataset2013hciold} and \textit{Inria DLFD} \cite{lfdataset2018inria}. The disparity range of the test dataset is much larger than that of light fields captured by commericial cameras, which is usually less than 1 pixel. It is worth noting that the baseline range between input corner views of a $7\times7$ light field are 6 times of the disparity range, i.e., in the range of [-24, 24].

        \begin{table*}[!t]
        \caption{Comparisons of running time (in second) of different methods.}
        \label{table_time}
        \centering
        \resizebox{0.85\textwidth}{!}{
        \begin{tabular}{c | c c c c | c }
        \toprule[2pt]
        Algorithms & Wu et al. \shortcite{lfrec2017wu:blur} & Wu et al. \shortcite{lfrec2019wu:shear} & Yeung et al. \shortcite{lfrec2018Yeung} & Kalantari et al. \shortcite{lfrec2016kalantari} &  Ours\\
        \midrule[1pt]
        running time &  257.70 & 101.70 & 0.48 & 168.86 & 3.49\\
        \bottomrule[2pt]
        \end{tabular}
        }
        \end{table*}
        
        \begin{figure*}[!t]
        \centering
        \includegraphics[width=0.84\textwidth]{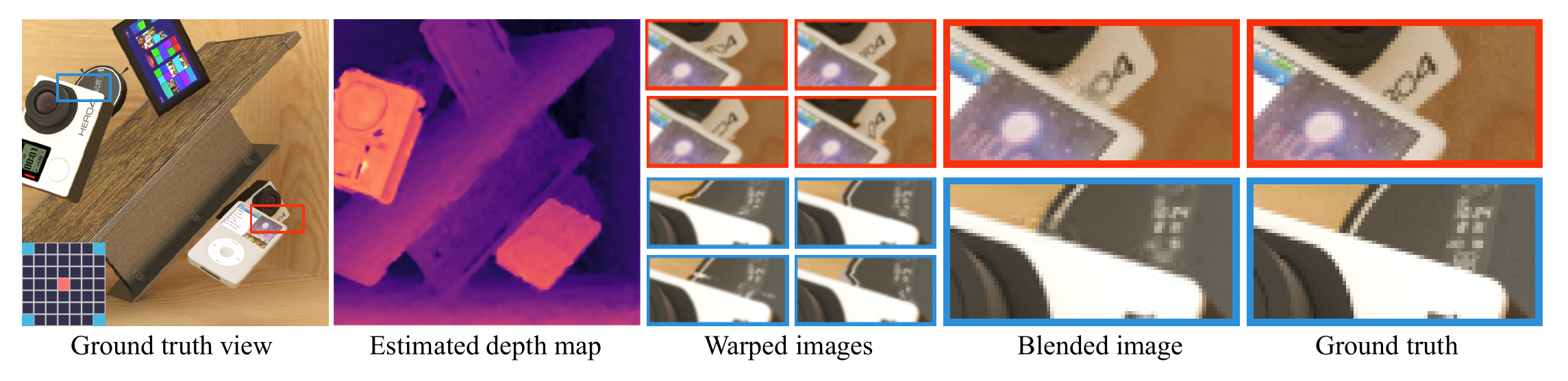}
        \caption{We qualitatively evaluated the effectiveness of our depth estimation and light field blending modules. The estimated depth map, and the zoomed-in images before and after the light field blending module are presented.}
        \label{fig_ablation}
        \end{figure*}
        
\textbf{Reconstruction evaluation.}
We used PSRN and SSIM to quantitatively evaluate all methods, and
Tables \ref{table_hci}, \ref{table_hci_old} and \ref{table_inria} list the results. We also presented the disparity range of each light field to investigate its effect on the reconstruction quality. It can be observed that the results of non-depth based method Wu et al. \shortcite{lfrec2017wu:blur} have very low PSNR (below 30dB) when the disparities of light fields are larger than 1.5 pixel. Although Yeung et al. \shortcite{lfrec2018Yeung} is able to achieve quiet good performance when the sampling baseline is relatively small (see the reuslts of \textit{buddha}, \textit{papillow} and \textit{monasroom} in Table. \ref{table_hci_old} ), the PSNR and SSIM of their results decrease greatly on light fields with a wider disparity range (see the results of \textit{stillLife} in Table \ref{table_hci_old} and \textit{dishes} in Table \ref{table_hci}). However, even under the help of depth information, Wu et al. \shortcite{lfrec2019wu:shear} and Kalantari et al. \shortcite{lfrec2016kalantari} only achieve performance comparable to Yeung et al. \shortcite{lfrec2018Yeung}, which indicates that the depth information is not fully utilized in these methods.
In contrast, the predictions of our method always have the highest quality, i.e., our method improves the PSNR around 1dB in small baseline sampling (disparities are smaller than 1.5 pixel) and more than 2dB in large baseline sampling (disparities are larger than 1.5 pixel),
which demonstrates the great advantages of our method.

We also provided visual comparisons of different methods, as shown in 
Fig. \ref{fig_visual_patch}. It can be seen that the predictions of Wu et al. \shortcite{lfrec2017wu:blur} have severe ghosting caused by the large disparity of objects, while different levels of artifacts appear  around occlusion boundaries in the results of other compared methods.
In contrast, our method produces high quality images which are closer to the ground truth ones.

\begin{figure}[!t]
        \centering
        \includegraphics[width=0.7\columnwidth]{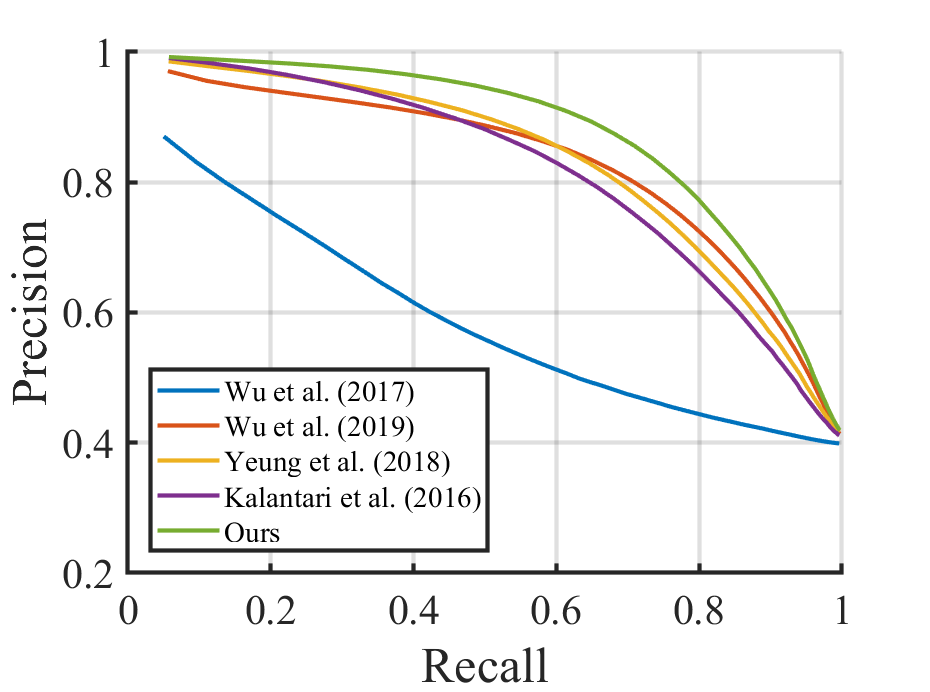}
        \caption{Comparisons of the parallax content PR curves for different methods.The PR curves are computed by averaging over all testing light field images.}
        \label{fig_prcurve}
        \end{figure}

To further evaluate the preservation of the light field parallax structure quantitatively, we compared the light field parallax edge precision-reall (PR) curves \cite{lfdenoise2018chen} of the angularly super-resolved light fields, and Fig. \ref{fig_prcurve} shows the results. It can be seen that the PR curve of our method is closer to the top-right corner compared with others, which shows that the light field structure is well maintained in the predictions of our method.

\textbf{Efficiency evaluation.}
Our method takes 3.49 seconds to reconstruct a $7\times7$ light field with spatial resolution of $512\times512$ from $2\times2$ views.
Specifically, it takes 0.39 seconds to initially synthesize all novel views and 3.50 seconds for light field blending. Table \ref{table_time} shows the comparisons of the running time for different methods to angularly super-resolve a light field. It can be seen that our method is significantly faster than other methods except Yeung et al. \shortcite{lfrec2018Yeung}. However, consider the compromise of Yeung et al. \shortcite{lfrec2018Yeung} on the performance, our method is superior. All methods were evaluated on a Intel 3.70 GHz desktop with 32 GB RAM and a GeForce RTX 2080 Ti GPU.

\textbf{Ablation study.}
In Fig. \ref{fig_ablation}, we demonstrated the effectiveness of three modules involved in our method. The estimated depth map for the center view, the warped images and the final reconstructed images after blending are presented. It can be observed that our depth estimation module performs good on most areas, but produce rough boundaries for some objects. Consequently, the warped images maintain sharp texture on plain areas, but have obvious distortions around occlusion boundaries.
Moreover, various artifacts appearing in the images warped from different input views are corrected in the blended images, which demonstrates the effectiveness of our light field blending module.

We also quantitatively compared the reconstruction quality of light field blending and view blending.
For fair comparisons, we built two networks using different blending strategies and the same depth estimation module, and trained them without the EPI gradient loss.
The results are listed in Table \ref{table_blend}, which demonstrates the advantage of our proposed light field blending.

        \begin{table}[!t]
        \caption{Quantitative comparisons (PSNR/SSIM) of view blending and light field blending.}
        \label{table_blend}
        \centering
        \resizebox{\columnwidth}{!}{
        \begin{tabular}{c | c c c }
        \toprule[2pt]
         & HCI & HCI old & Inria DLFD\\
        \midrule[1pt]
        View blending & 32.69/0.920 & 39.25/0.954 & 33.17/0.920 \\
        Light field blending & 34.29/0.933 & 40.70/0.958 & 34.56/0.929 \\
        \bottomrule[2pt]
        \end{tabular}
        }
        \end{table}

\section{Conclusion}
We have presented a learning-based method for light field angular super-resolution. More precisely, we focused on the reconstruction of a high angular resolution light field from a small set of input views with a large baseline. By explicitly modeling the scene geometry for novel view synthesis and efficiently exploring the angular relations for light field blending, our method outperforms the state-of-the-art ones on the task of super-resolving light fields of angular resolution $2\times 2$ to those of angular resolution $7\times7$  over various datasets with a disparity range of $[-4,4]$, i.e., our method improves the PSNR of the second best method up to 2 dB in average, while saves the execution time 48$\times$. In addition, our method preserves the light field parallax  structure better.

\bibliographystyle{aaai}
\bibliography{references}        

\begin{thebibliography}{}

\bibitem[\protect\citeauthoryear{{Babacan} \bgroup et al\mbox.\egroup
  }{2012}]{lfrec2012compressive}
{Babacan}, S.~D.; {Ansorge}, R.; {Luessi}, M.; {Mataran}, P.~R.; {Molina}, R.;
  and {Katsaggelos}, A.~K.
\newblock 2012.
\newblock Compressive light field sensing.
\newblock {\em IEEE Transactions on Image Processing} 21(12):4746--4757.

\bibitem[\protect\citeauthoryear{Chen, Hou, and Chau}{2018}]{lfdenoise2018chen}
Chen, J.; Hou, J.; and Chau, L.-P.
\newblock 2018.
\newblock Light field denoising via anisotropic parallax analysis in a cnn
  framework.
\newblock {\em IEEE Signal Processing Letters} 25(9):1403--1407.

\bibitem[\protect\citeauthoryear{Dong \bgroup et al\mbox.\egroup
  }{2014}]{sisr2014srcnn}
Dong, C.; Loy, C.~C.; He, K.; and Tang, X.
\newblock 2014.
\newblock Learning a deep convolutional network for image super-resolution.
\newblock In {\em European Conference on Computer Vision (ECCV)},  184--199.

\bibitem[\protect\citeauthoryear{{Gupta} \bgroup et al\mbox.\egroup
  }{2017}]{lfrec2017compressiveDeep}
{Gupta}, M.; {Jauhari}, A.; {Kulkarni}, K.; {Jayasuriya}, S.; {Molnar}, A.; and
  {Turaga}, P.
\newblock 2017.
\newblock Compressive light field reconstructions using deep learning.
\newblock In {\em IEEE Conference on Computer Vision and Pattern Recognition
  Workshops (CVPRW)},  1277--1286.

\bibitem[\protect\citeauthoryear{Honauer \bgroup et al\mbox.\egroup
  }{2016}]{lfdataset2016hci}
Honauer, K.; Johannsen, O.; Kondermann, D.; and Goldluecke, B.
\newblock 2016.
\newblock A dataset and evaluation methodology for depth estimation on 4d light
  fields.
\newblock In {\em Asian Conference on Computer Vision (ACCV)},  19--34.

\bibitem[\protect\citeauthoryear{Jeon \bgroup et al\mbox.\egroup
  }{2015}]{lfdepth2015jeon}
Jeon, H.-G.; Park, J.; Choe, G.; Park, J.; Bok, Y.; Tai, Y.-W.; and So~Kweon,
  I.
\newblock 2015.
\newblock Accurate depth map estimation from a lenslet light field camera.
\newblock In {\em IEEE Conference on Computer Vision and Pattern Recognition
  (CVPR)},  1547--1555.

\bibitem[\protect\citeauthoryear{Kalantari, Wang, and
  Ramamoorthi}{2016}]{lfrec2016kalantari}
Kalantari, N.~K.; Wang, T.-C.; and Ramamoorthi, R.
\newblock 2016.
\newblock Learning-based view synthesis for light field cameras.
\newblock {\em ACM Transactions on Graphics} 35(6):193:1--193:10.

\bibitem[\protect\citeauthoryear{Kamal \bgroup et al\mbox.\egroup
  }{2016}]{lfrec2016tensor}
Kamal, M.~H.; Heshmat, B.; Raskar, R.; Vandergheynst, P.; and Wetzstein, G.
\newblock 2016.
\newblock Tensor low-rank and sparse light field photography.
\newblock {\em Computer Vision Image Understanding} 145(C):172--181.

\bibitem[\protect\citeauthoryear{Kingma and Ba}{2014}]{kingma2014adam}
Kingma, D.~P., and Ba, J.
\newblock 2014.
\newblock Adam: A method for stochastic optimization.
\newblock {\em arXiv preprint arXiv:1412.6980}.

\bibitem[\protect\citeauthoryear{Krizhevsky, Sutskever, and
  Hinton}{2012}]{krizhevsky2012classification}
Krizhevsky, A.; Sutskever, I.; and Hinton, G.~E.
\newblock 2012.
\newblock Imagenet classification with deep convolutional neural networks.
\newblock In {\em Advances in neural information processing systems (NeurIPS)},
   1097--1105.

\bibitem[\protect\citeauthoryear{Levin, Freeman, and
  Durand}{2008}]{lfrec2008bayesian}
Levin, A.; Freeman, W.~T.; and Durand, F.
\newblock 2008.
\newblock Understanding camera trade-offs through a bayesian analysis of light
  field projections.
\newblock In {\em European Conference on Computer Vision (ECCV)},  88--101.

\bibitem[\protect\citeauthoryear{Lytro}{2016}]{lytro}
Lytro.
\newblock 2016.
\newblock \url{https://www.lytro.com/}.
\newblock [Online].

\bibitem[\protect\citeauthoryear{Marwah \bgroup et al\mbox.\egroup
  }{2013}]{lfrec2013compressive}
Marwah, K.; Wetzstein, G.; Bando, Y.; and Raskar, R.
\newblock 2013.
\newblock Compressive light field photography using overcomplete dictionaries
  and optimized projections.
\newblock {\em ACM Transactions on Graphics} 32(4):46:1--46:12.

\bibitem[\protect\citeauthoryear{Mildenhall \bgroup et al\mbox.\egroup
  }{2019}]{viewsyn2019localfusion}
Mildenhall, B.; Srinivasan, P.~P.; Ortiz-Cayon, R.; Kalantari, N.~K.;
  Ramamoorthi, R.; Ng, R.; and Kar, A.
\newblock 2019.
\newblock Local light field fusion: Practical view synthesis with prescriptive
  sampling guidelines.
\newblock {\em ACM Transactions on Graphics} 38(4):29:1--29:14.

\bibitem[\protect\citeauthoryear{{Mitra} and
  {Veeraraghavan}}{2012}]{lfrec2012gmm}
{Mitra}, K., and {Veeraraghavan}, A.
\newblock 2012.
\newblock Light field denoising, light field superresolution and stereo camera
  based refocussing using a gmm light field patch prior.
\newblock In {\em IEEE Conference on Computer Vision and Pattern Recognition
  Workshops (CVPRW)},  22--28.

\bibitem[\protect\citeauthoryear{Niklaus, Mai, and
  Liu}{2017}]{sepfilter2017video}
Niklaus, S.; Mai, L.; and Liu, F.
\newblock 2017.
\newblock Video frame interpolation via adaptive separable convolution.
\newblock In {\em IEEE International Conference on Computer Vision (ICCV)},
  261--270.

\bibitem[\protect\citeauthoryear{Raytrix}{2016}]{raytrix}
Raytrix.
\newblock 2016.
\newblock \url{https://www.raytrix.de/}.
\newblock [Online].

\bibitem[\protect\citeauthoryear{Shi \bgroup et al\mbox.\egroup
  }{2014}]{lfrec2014sparsity}
Shi, L.; Hassanieh, H.; Davis, A.; Katabi, D.; and Durand, F.
\newblock 2014.
\newblock Light field reconstruction using sparsity in the continuous fourier
  domain.
\newblock {\em ACM Transactions on Graphics} 34(1):12:1--12:13.

\bibitem[\protect\citeauthoryear{Shi, Jiang, and
  Guillemot}{2019}]{lfdataset2018inria}
Shi, J.; Jiang, X.; and Guillemot, C.
\newblock 2019.
\newblock A framework for learning depth from a flexible subset of dense and
  sparse light field views.
\newblock {\em IEEE Transactions on Image Processing}  1--15.

\bibitem[\protect\citeauthoryear{{Vagharshakyan}, {Bregovic}, and
  {Gotchev}}{2018}]{lfrec2018shearlet}
{Vagharshakyan}, S.; {Bregovic}, R.; and {Gotchev}, A.
\newblock 2018.
\newblock Light field reconstruction using shearlet transform.
\newblock {\em IEEE Transactions on Pattern Analysis and Machine Intelligence}
  40(1):133--147.

\bibitem[\protect\citeauthoryear{Vaish and Adams}{2008}]{lfgantry}
Vaish, V., and Adams, A.
\newblock 2008.
\newblock {The (New) Stanford Light Field Archive}.
\newblock \url{http://lightfield.stanford.edu/acq.html}.
\newblock [Online].

\bibitem[\protect\citeauthoryear{Wang \bgroup et al\mbox.\egroup
  }{2018a}]{flow2018occlusion}
Wang, Y.; Yang, Y.; Yang, Z.; Zhao, L.; Wang, P.; and Xu, W.
\newblock 2018a.
\newblock Occlusion aware unsupervised learning of optical flow.
\newblock In {\em IEEE Conference on Computer Vision and Pattern Recognition
  (CVPR)},  4884--4893.

\bibitem[\protect\citeauthoryear{Wang \bgroup et al\mbox.\egroup
  }{2018b}]{lfrec2018pseudo}
Wang, Y.; Liu, F.; Wang, Z.; Hou, G.; Sun, Z.; and Tan, T.
\newblock 2018b.
\newblock End-to-end view synthesis for light field imaging with pseudo 4dcnn.
\newblock In {\em European Conference on Computer Vision (ECCV)},  333--348.

\bibitem[\protect\citeauthoryear{{Wanner} and
  {Goldluecke}}{2014}]{lfdepth2014wanner}
{Wanner}, S., and {Goldluecke}, B.
\newblock 2014.
\newblock Variational light field analysis for disparity estimation and
  super-resolution.
\newblock {\em IEEE Transactions on Pattern Analysis and Machine Intelligence}
  36(3):606--619.

\bibitem[\protect\citeauthoryear{Wanner, Meister, and
  Goldluecke}{2013}]{lfdataset2013hciold}
Wanner, S.; Meister, S.; and Goldluecke, B.
\newblock 2013.
\newblock Datasets and benchmarks for densely sampled 4d light fields.
\newblock In {\em VMV},  225--226.

\bibitem[\protect\citeauthoryear{Wilburn \bgroup et al\mbox.\egroup
  }{2005}]{lf2005array}
Wilburn, B.; Joshi, N.; Vaish, V.; Talvala, E.-V.; Antunez, E.; Barth, A.;
  Adams, A.; Horowitz, M.; and Levoy, M.
\newblock 2005.
\newblock High performance imaging using large camera arrays.
\newblock {\em ACM Transaction on Graphics} 24(3):765--776.

\bibitem[\protect\citeauthoryear{{Wu} \bgroup et al\mbox.\egroup
  }{2017}]{lfrec2017wu:blur}
{Wu}, G.; {Zhao}, M.; {Wang}, L.; {Dai}, Q.; {Chai}, T.; and {Liu}, Y.
\newblock 2017.
\newblock Light field reconstruction using deep convolutional network on epi.
\newblock In {\em IEEE Conference on Computer Vision and Pattern Recognition
  (CVPR)},  1638--1646.

\bibitem[\protect\citeauthoryear{{Wu} \bgroup et al\mbox.\egroup
  }{2019}]{lfrec2019wu:shear}
{Wu}, G.; {Liu}, Y.; {Dai}, Q.; and {Chai}, T.
\newblock 2019.
\newblock Learning sheared epi structure for light field reconstruction.
\newblock {\em IEEE Transactions on Image Processing} 28(7):3261--3273.

\bibitem[\protect\citeauthoryear{Yeung \bgroup et al\mbox.\egroup
  }{2018}]{lfrec2018Yeung}
Yeung, W. F.~H.; Hou, J.; Chen, J.; Chung, Y.~Y.; and Chen, X.
\newblock 2018.
\newblock Fast light field reconstruction with deep coarse-to-fine modeling of
  spatial-angular clues.
\newblock In {\em European Conference on Computer Vision (ECCV)},  137--152.

\bibitem[\protect\citeauthoryear{Yeung \bgroup et al\mbox.\egroup
  }{2019}]{lfsr2019sas}
Yeung, H. W.~F.; Hou, J.; Chen, X.; Chen, J.; Chen, Z.; and Chung, Y.~Y.
\newblock 2019.
\newblock Light field spatial super-resolution using deep efficient
  spatial-angular separable convolution.
\newblock {\em IEEE Transactions on Image Processing} 28(5):2319--2330.

\bibitem[\protect\citeauthoryear{Yoon \bgroup et al\mbox.\egroup
  }{2015}]{lfrec2015yoon}
Yoon, Y.; Jeon, H.-G.; Yoo, D.; Lee, J.-Y.; and So~Kweon, I.
\newblock 2015.
\newblock Learning a deep convolutional network for light-field image
  super-resolution.
\newblock In {\em IEEE International Conference on Computer Vision Workshops
  (ICCVW)},  24--32.

\bibitem[\protect\citeauthoryear{Zhou \bgroup et al\mbox.\egroup
  }{2018}]{viewsyn2018mpi}
Zhou, T.; Tucker, R.; Flynn, J.; Fyffe, G.; and Snavely, N.
\newblock 2018.
\newblock Stereo magnification: Learning view synthesis using multiplane
  images.
\newblock {\em ACM Transactions on Graphics} 37(4):65:1--65:12.

\end{thebibliography}
\end{document}